\documentclass[preprint,10.5pt]{elsarticle}
\usepackage[margin=1in]{geometry}
\pdfoutput=1 

\usepackage[pdftex,bookmarksnumbered]{hyperref}
\usepackage[mode=buildnew]{standalone}
\usepackage{amssymb,amsmath,amsthm,amsbsy}
\usepackage{systeme}
\usepackage{physics}
\usepackage{hyperref} 
\usepackage{graphicx}
\usepackage{epstopdf}
\usepackage{epsfig}
\usepackage{multicol}
\usepackage{subfigure}
\usepackage{enumitem}
\usepackage{multirow}
\usepackage{algorithm}
\usepackage{algpseudocode}
\usepackage{bm}
\usepackage{cleveref}
\usepackage{setspace}
\usepackage{color}
\usepackage{todonotes}
\usepackage[mathlines,running]{lineno}
\usepackage{booktabs}
\usepackage{multirow}
\usepackage{siunitx}
\usepackage{longtable}
\usepackage{siunitx,upgreek}

\usepackage{amssymb,amsmath,amsthm,amsbsy}

\newcommand*{\defeq}{\mathrel{\vcenter{\baselineskip0.5ex \lineskiplimit0pt
                     \hbox{\scriptsize.}\hbox{\scriptsize.}}}%
                     =}
\newcommand*{\defeqinv}{=\mathrel{\vcenter{\baselineskip0.5ex \lineskiplimit0pt
                     \hbox{\scriptsize.}\hbox{\scriptsize.}}}}

\newcommand{\encoder}[1]{\Phi^{(#1)}}
\newcommand{\decoder}[1]{\Psi^{(#1)}}
\newcommand{\x}[1]{{\mathbf{x}}^{(#1)}}
\newcommand{\h}[1]{{\mathbf{h}}^{(#1)}}
\newcommand{\htot}[1]{{\mathbf{h}}_\text{tot}^{(#1)}}

\newcommand{\yhat}[1]{{\hat{\mathbf{y}}}^{(#1)}}
\newcommand{\R}{\mathbb{R}}
\newcommand{\W}{\mathbf{W}}

\newcommand{\paolo}[1]{\textcolor{black}{#1}}

\begin{document}
\makeatletter
\def\ps@pprintTitle{%
  \let\@oddhead\@empty
  \let\@evenhead\@empty
  \let\@oddfoot\@empty
  \let\@evenfoot\@oddfoot
}
\makeatother


\begin{frontmatter}

\title{Progressive multi-fidelity learning {with neural networks} \\for physical system predictions}

\author[add1,add4]{Paolo Conti\corref{cor1}}
\ead{pconti@turing.ac.uk}
\cortext[cor1]{Corresponding author.}

\author[add2]{Mengwu Guo}
\ead{mengwu.guo@math.lu.se}

\author[add3]{Attilio Frangi}
\ead{attilio.frangi@polimi.it}

\author[add4]{Andrea Manzoni}
\ead{andrea1.manzoni@polimi.it}

\address[add1]{The Alan Turing Institute, London, NW1 2DB, UK}
\address[add2]{Centre for Mathematical Sciences, Lund University, Sweden}
\address[add3]{Department of Civil and Environmental Engineering, Politecnico di Milano, Italy}
\address[add4]{MOX -- Department of Mathematics, Politecnico di Milano, Italy}

\begin{abstract}
Highly accurate datasets from numerical or physical experiments are often expensive and time-consuming to acquire, posing a significant challenge for applications that require precise evaluations, potentially across multiple scenarios and in real-time. Even building sufficiently accurate surrogate models can be extremely challenging with limited high-fidelity data. Conversely, less expensive, low-fidelity data can be computed more easily and encompass a broader range of scenarios. By leveraging multi-fidelity information, prediction capabilities of surrogates can be improved.
However, in practical situations, data may be different in types, come from sources of different modalities, and not be concurrently available, further complicating the modeling process. 
To address these challenges, we introduce a progressive multi-fidelity surrogate model. This model can sequentially incorporate diverse data types using tailored encoders. Multi-fidelity regression from the encoded inputs to the target quantities of interest is then performed using neural networks. Input information progressively flows from lower to higher fidelity levels through two sets of connections: \textit{concatenations} among all the encoded inputs, and \textit{additive connections} among the final outputs. This dual connection system enables the model to exploit correlations among different datasets while ensuring that each level makes an additive correction to the previous level without altering it. This approach prevents performance degradation as new input data are integrated into the model and automatically adapts predictions based on the available inputs.
We demonstrate the effectiveness of the approach on numerical benchmarks and a real-world air pollution case study, showing that it reliably integrates multi-modal data, mitigates low-fidelity imperfections, and provides accurate predictions, while maintaining performance when generalizing across time and parameter variations.


\end{abstract}


\end{frontmatter}

\section{Introduction}
Scientific computing today stands as a cornerstone in several fields, delivering unparalleled success in diverse applications in science and engineering, including weather forecast \cite{price2025probabilistic, bodnar2025foundation, allen2025end}, computational biology \cite{jumper2021highly}, among many others. 
However, in spite of its undeniable success, the computational demands of high-fidelity simulations may become prohibitively expensive. 
This computational burden becomes critical in multi-query scenarios such as uncertainty quantification, optimal control, or model calibration, where repeated evaluations are required. 
In these challenging scenarios, the construction of efficient surrogate models thus emerges as a crucial strategy. 
Such models serve as indispensable proxies, enabling the cost-effective and accurate prediction of physical systems.
In particular, data-driven surrogates are capable of learning complex system dynamics directly from high-dimensional datasets, without relying on explicit physical knowledge of the underlying process.
By extracting rich low-dimensional latent representations and capturing nonlinear correlations, these models provide accurate reduced-order approximations, making them a powerful tool for efficient and scalable scientific computing \cite{QMN, benner2005dimension, Noack2011book, HRS, Lee2020autoencoder, fresca2020comprehensive}.

However, the effectiveness and trustworthiness of deep learning surrogates can deteriorate when acquiring high-fidelity data is prohibitively expensive to span the entire modeling domain adequately or even to fit a surrogate model. 
On the other hand, we might frequently rely on a wider range of low-fidelity data, obtained for instance from coarser simulations or simplified models, as well as heterogeneous sources of information.
These additional datasets may differ in nature and modality, carrying complementary information: effectively integrating multiple data streams has the potential to improve predictive capabilities and foster a more holistic understanding of complex physical systems.

In these scenarios, multi-fidelity strategies emerge as a powerful solution in many areas of science and engineering \cite{peherstorfer2018survey, mfPOD, ahmed2021multifidelity, kast2020non, geneva2020multi}.
By leveraging data of different fidelity levels, multi-fidelity approaches enhance the model's ability to generalize across different regions, particularly where high-fidelity information is limited or unavailable, thereby achieving an effective compromise between computational cost and model accuracy.
{A wide range of multi-fidelity regression methods have been developed using Gaussian processes \cite{Hagan2000, perdikaris2017nonlinear, alvarez2012kernels, lam2015multifidelity}, as well as other classical surrogate modeling techniques, including polynomial chaos expansions \cite{ng2012multifidelity,palar2016multi}.
More recently, deep learning has been proposed to address challenges related to high dimensionality, scalability, and strongly nonlinear cross-fidelity correlations \cite{meng2020composite, liu2019multi, mfLSTM, guo2022multi, Motamed}. Within this line of work, several learning strategies have been explored for multi-fidelity data fusion, including heterogeneous domain training \cite{sarkar2019multi}, manifold alignment \cite{perron2021multi}, diffusion models \cite{shi2025diffusion}, and reinforcement learning approaches \cite{cutler2014reinforcement,sun2025multi}.}
{Nevertheless, many existing approaches are tailored to settings in which datasets are structurally homogeneous. In this context, different fidelity levels provide information about the same quantity of interest, obtained from models or estimators of varying accuracy or computational cost. 
Typical examples include solution data generated using the same numerical solver at different resolution levels, or data produced by different solvers targeting the same output fields \cite{mfPOD}.}
Therefore, challenges arise in incorporating diverse sources of information or different data types, a problem often referred to in machine learning as \textit{multi-modal} learning, impacting a wealth of application fields ranging from
cancer biomarker discovery \cite{steyaert2023multimodal}, precision medicine \cite{boehm2022harnessing} and neuroimaging \cite{zhang2020advances} to 
wearable sensors \cite{narayanswamy2024scaling} and meteorological monitoring \cite{lahat2015multimodal}, just to mention a few.
Additionally, datasets may not be available simultaneously at the time of surrogate construction but acquired at different stages, in which case retraining neural networks with new inputs may lead to the issue known as \textit{catastrophic forgetting}.

To tackle these challenges, we propose a progressive multi-fidelity surrogate modeling paradigm that incorporates new datasets of varying fidelity as they become available.
This allows for progressively enhancing prediction accuracy and reducing uncertainty, while effectively ensuring knowledge retention across updates.
This model can sequentially incorporate diverse data types using tailored encoder neural networks. 
The encoders process and transform data into a meaningful latent representation that is suitable to be merged with the other fidelity datasets. 
The architecture of each encoder is selected according to the data modality: for example, feedforward networks for vector-valued data, recurrent networks (e.g., LSTMs \cite{graves2012long}) for time series, convolutional networks \cite{lecun2002gradient} or vision transformers \cite{dosovitskiy2020image} for image data, and so on.
The encoded features are concatenated and mapped to the target quantities of interest through a decoder, which is trained in a supervised manner on high-fidelity data.

\begin{figure}[t]
    \centering
    \includegraphics[width=.9\linewidth]{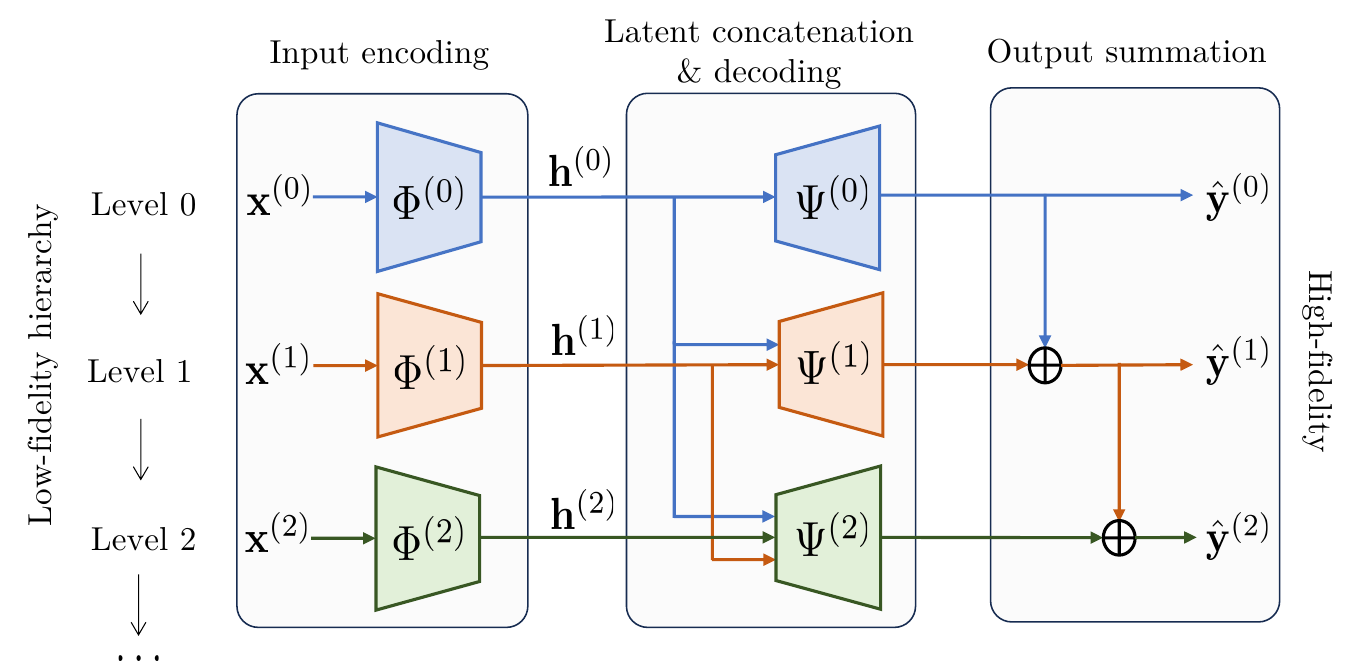}
    \caption{Progressive multi-fidelity surrogate modeling structure. The method is composed of three steps: (i) \textit{input encoding}, at each level $l$, input data $\x{l}$ are mapped by encoders $\encoder{l}$ into latent variables $\h{l}$; (ii) \textit{latent concatenation and decoding}, latent variables are concatenated and decoded by $\decoder{l}$; (iii) \textit{output summation}, previous output is summed to the current one to form the final output $\yhat{l}$.}
    \label{fig: method}
\end{figure}

Fig.~\ref{fig: method} illustrates how input information flows from lower to higher fidelity levels through two types of connections: \textit{concatenations} among all encoded inputs and \textit{additive connections} among the decoded outputs. This dual connection system enables exploiting correlations among different datasets while ensuring that each level makes an additive correction to the previous level without altering it. This approach protects against catastrophic forgetting and performance degradation as new input data is integrated into the model. 


Moreover, this progressive multi-fidelity structure is beneficial not only during training but also in online deployment.  {Since each fidelity level contributes an additive correction, the model can operate with only the input modalities available at inference time, by construction. 
When higher-fidelity inputs are unavailable due to time, budget, or data acquisition constraints, the framework systematically falls back to lower-fidelity levels.
This enables automatic adaptation to asynchronous or missing inputs, ensuring operational continuity and progressively refined predictions as additional data become available.}
By leveraging the inductive bias of the progressive structure and the information contained in physically meaningful low-fidelity data, the framework mitigates the lack of physical consistency typical of data-driven methods in data-scarce regimes, thereby enhancing accuracy in generalization across time and parameters and improving reliability. 

To evaluate the framework, we consider three applications spanning both {\em (i)} synthetic data obtained through the solution of nonlinear, time-dependent partial differential equations (PDEs) and {\em (ii)} real-world data. 
First, a reaction-diffusion system reconstructs high-fidelity spiral wave dynamics from coarse, noisy low-fidelity simulations. 
Second, a Navier-Stokes benchmark reconstructs vortex shedding from hierarchical low-fidelity inputs, including drag/lift coefficients, outlet sensors, and partial-domain snapshots. 
Third, an air pollution scenario estimates benzene concentrations from low-cost sensor data despite missing or unreliable inputs. 
These cases highlight key challenges: integrating multi-modal data, mitigating low-fidelity imperfections, and enabling reliable temporal and parametric extrapolation.

Our method builds upon the progressive networks proposed by \cite{rusu2016progressive}, which employ lateral connections to accumulate knowledge across reinforcement learning tasks. We generalize this concept to the context of multi-fidelity and multi-modal learning for physical systems. Related approaches include multi-fidelity neural processes \cite{wu2022multi}, which hierarchically integrate datasets but do not incorporate additive corrections to prevent catastrophic forgetting, and multi-fidelity reduced-order surrogate models \cite{mfPOD}, which are limited to two fidelity levels of homogeneous modality.

In summary, our contributions include:
    \begin{itemize}
    \item A progressive multi-fidelity architecture that sequentially integrates heterogeneous datasets while preserving prior knowledge through additive corrections.
    \item A dual fusion strategy that integrates multi-fidelity data both at the latent level (early fusion of encoded features) and at the output level (late fusion via output corrections), enabling effective exploitation of complementary information across modalities.
    \item Extensive validation across a range of PDE benchmarks with diverse quantities of interest, as well as a real-world air pollution case study, demonstrating the flexibility of the proposed architecture and practical relevance in cases where noisy data are acquired from real sensor measurements.
\end{itemize}

The paper is structured as follows. 
In Section~\ref{sec: method}, we detail the structure of the progressive multi-fidelity learning model and describe the procedures for offline training and online deployment.
Sections~\ref{sec: RD}–\ref{sec: NS}–\ref{sec: air_pollution} present and discuss the model’s performance on the three applications introduced earlier, with conclusions drawn in Section~\ref{sect: conclusions}. 
The source code of the proposed method, along with the datasets used in this work, is available in the public repository \url{https://github.com/ContiPaolo/Progressive-Multifidelity-NNs}.

\section{Method}
\label{sec: method}

In this section, we present the method for constructing the progressive multi-fidelity surrogate model for physical system predictions.
We assume that the low-fidelity input data can be arranged in a hierarchy, and, for simplicity, that there is a single high-fidelity output to be predicted.
In the following, we detail the offline training and the online deployment phases of the method.

\subsection{Offline training}
\label{sect: training}
The progressive multi-fidelity surrogate model is constructed through sequential levels, where at each level, a different low-fidelity dataset is provided as input, following the hierarchy.
Each level is characterized by three steps: input encoding, decoding of latent concatenation, and output summation. Figure~\ref{fig: method} presents a schematic overview of the approach.

\begin{itemize}
    \item \textbf{Input encoding}. Since data sources can be of any kind, inputs are processed by an encoder neural network and transformed into a latent representation that is suitable to be merged with the other fidelity datasets. The choice of the encoder network architecture reflects the input data type. For instance, standard feedforward (FF) neural networks (NNs) can be employed with parameter values and vector-valued data; recurrent NNs, such as long short-term memory (LSTM) networks, for time series; convolutional NNs or vision transformers for image data. Hence, at each level $l$, the input data $\x{l} \in \R^{d_{\text{in}}^{(l)}}$ are passed to an encoder network $\encoder{l}$:
    \begin{equation}
         \encoder{l}(\cdot\,; {\W}_ \encoder{l}):\R^{d_\text{in}^{(l)}}\rightarrow \mathbb{R}^{d_\text{h}^{(l)}}, \;\text{such that}\; {\x{l}} \mapsto \encoder{l}(\x{l})\defeqinv\h{l},
    \end{equation}
     where $\h{l}$ denotes the latent variable of dimension $d_\text{h}^{(l)}$, ${\W}_ \encoder{l}$ the network parameters and $d_\text{in}^{(l)}$ indicates the original dimension of the input data. 
     The data encoding process is crucial for extracting data features and providing low-dimensional latent representation, especially when dealing with high-dimensional data by considering $d_\text{h}^{(l)} \ll d_\text{in}^{(l)}$. 
     As data types might be different across different levels, the encoding process is specific to each level and independent from the others. Moreover, it is not necessary to use a neural network as the encoder; other encoding techniques allowing for dimensionality reduction or change of coordinates, such as Proper Orthogonal Decomposition (POD), can also be considered, as already proposed in \cite{mfPOD}.
     
     \item \textbf{Latent concatenation and decoding}. The current latent variable $\h{l}$ is concatenated with all latent variables from previous levels $\{\h{j}\}_{j<l}$ and processed by a decoder neural network. The role of the decoder is to map the concatenated latent variables into the output quantity of interest we aim to estimate. We expect the decoder to merge all fidelity information and exploit the correlations among all datasets. The type of the decoder network reflects the output data type and it remains the same across all levels, as each level aims to estimate the same high-fidelity output of interest. For instance, in physical systems governed by PDEs, spatio-temporal decoders are employed to reconstruct the evolution in time of the solution field. 
    The decoder at level $l$ is denoted as $\decoder{l}$ and defined as:
    \begin{equation}
         \decoder{l}(\cdot\,; {\W}_ \decoder{l}):\R^{d_{\text{h}_\text{tot}}^{(l)}}\rightarrow \mathbb{R}^{d_\text{out}}, \;\text{such that}\; \htot{l} \mapsto \decoder{l}(\htot{l}),
    \end{equation}
    where $\htot{l} \defeq [\h{0} \cdots \h{l-1} \h{l}]\in \R^{d_{\text{h}_\text{tot}}^{(l)}}$ with ${d_{\text{h}_\text{tot}}^{(l)}} = \sum_{j\leq l} d^{(j)}_\text{h}$, which represents the dimension of the concatenated latent variables that serve as input to the decoder. ${\W}_\decoder{l}$  are the decoder network parameters. The output dimension of the decoder $d_\text{out}$ is the same at every level. Note that the present framework can be straightforwardly extended for the prediction of multiple outputs by considering multiple decoders.
    
    \item \textbf{Output summation}. In order to preserve information across the different levels, the output of the decoder is summed to the predicted output at the preceding level. Except for the first level, where the final output simply coincides with the decoder output $\yhat{0} = \decoder{0}(\h{0})$, for all $l>0$ the final output is defined as
    \begin{equation}
    \yhat{l} = \yhat{l-1} + \decoder{l}(\htot{l}){.}
    \end{equation}
    In this way, the current level contributes an additive correction approximating the residual between the ground truth and the previous estimate, while avoiding the loss of information contained in predictions from earlier levels.
\end{itemize}

The three steps outlined above define how input data are processed, at each level, to produce a set of progressively refined estimates $\{\yhat{l}\}_{l=0:L}$, approximating the high-fidelity target quantities of interest $\mathbf{y}$.
At the $l$-th level, the optimal parameters ${\W}_\encoder{j} \text{ and } {\W}_ \decoder{j}$ of the neural networks for encoding and decoding, $\encoder{l}$ and $\decoder{l}$ respectively, are obtained by solving the following optimization problem through back-propagation with {the} ADAM \cite{adam} algorithm:
\begin{equation}
\min_{{\W}_\encoder{j}, {\W}_ \decoder{j}}\; \frac{1}{N}\sum_{n=1}^{N}
\norm{{\hat{\mathbf{y}}}^{(l)}_n  - \mathbf{y}^{(l)}_n}^2_{2} + \lambda_\text{reg}\left({\W}_\encoder{j}, {\W}_ \decoder{j}\right),
\label{eq: loss}
\end{equation}    
where $\hat{\mathbf{y}}^{(l)}_n = \hat{\mathbf{y}}^{(l)}_n \left( \{\mathbf{x}_n^{(j)}\}_{j\leq l}\,;\{{\W}_\encoder{j}, {\W}_ \decoder{j}\}_{j\leq l} \right)$, and $n=1,\dots,N$ are the training data samples. \\
{This loss function is defined as the mean squared error (MSE) between the predicted output at level $l$ and the high-fidelity data, augmented with an additional regularization term $\lambda_\text{reg}$ to mitigate overfitting, which is particularly important given the limited amount of high-fidelity training data. 
In this work, $L_2$ regularization is employed (rather than alternatives such as $L_1$ or dropout), as it provides a smooth penalty on the network weights and is well suited for regression tasks involving real-valued quantities.
In particular, the regularization term is defined as
\begin{equation}
\lambda_\text{reg}\left({\W}_\encoder{j}, {\W}_ \decoder{j}\right)
= {\lambda_{\Phi}}\norm{{\W}_\encoder{j}}_2^2
+ {\lambda_{\Psi}}\norm{{\W}_\decoder{j}}_2^2,
\qquad \lambda_\Phi ,\lambda_\Psi \in \mathbb{R}^+ .
\label{eq: loss_reg}
\end{equation}
}
The different levels are trained sequentially, following the hierarchy order. In particular, at the $l$-th level, the optimization process described in equation \eqref{eq: loss} determines only the current encoder and decoder networks, while the network weights $\{\W_\Phi^{(j)}, \W_\Psi^{(j)}\}_{j<l}$ remain fixed. This practice ensures that at each new level, the structure at lower levels remains unaltered during the optimization process, thus preventing the occurrence of catastrophic forgetting, which could otherwise degrade the model’s performance at previous hierarchical levels.

The preceding levels convey information to the subsequent ones through both the concatenation of latent variables and the additive feedback from the previous output. Consequently, each new level configures itself as a correction to be applied to the previous level, which, in contrast, remains unchanged. {This additive formulation, together with freezing all previously trained components, enforces a one-directional flow of information, preserving earlier representations while higher-fidelity data contribute only via a learnable correction.}
{In this sense, latent concatenation implements early fusion by integrating multi-fidelity inputs into a shared latent space, whereas additive corrections act as late fusion, progressively refining the output. This combination is expected to provide the benefits of both fusion strategies in the data integration process.}

To ensure robustness and reliability of the predictions, the proposed method is equipped with uncertainty quantification.
In this work, uncertainty in the model predictions is estimated using an ensemble-based technique \cite{dietterich2000ensemble}, where the network is retrained multiple times with randomly initialized weights, and the statistical moments of the resulting ensemble of predictions are computed.
As a potential future development, the decoder could be replaced by neural network architectures that inherently provide uncertainty estimates in a more computationally efficient manner, such as Bayesian NNs \cite{mackay1995probable}, Deep Kernel Learning methods \cite{wilson2016deep}, or Conditional Neural Processes \cite{garnelo2018conditional}.

\subsection{Online prediction}
\begin{algorithm}[t]
\caption{Online Deployment of Progressive Multi-Fidelity Surrogate Model}
\label{alg:online}
\begin{algorithmic}[1]
\State \textbf{Input:} Trained encoders $\{\encoder{j}\}_{j=0}^{L}$, trained decoders $\{\decoder{j}\}_{j=0}^{L}$, available low-fidelity inputs $\{\mathbf{x}^{(j)}\}_{j=0}^{\bar{l}}$ up to level $\bar{l} \leq L$
\State \textbf{Output:} Predicted high-fidelity output $\hat{\mathbf{y}}^{(\bar{l})}$

\State Initialize prediction: $\hat{\mathbf{y}}^{(-1)} \gets \mathbf{0}$
\State Initialize latent concatenation: $\htot{-1} \gets \emptyset$
\For{$l = 0 \;\textbf{to}\; \bar{l}$}
    \State \textbf{Encode input}: $\h{l} \gets \encoder{l}(\x{l})$
    \State \textbf{Concatenate latents}: $\htot{l} \gets [\htot{l-1}; \h{l}]$
    \State \textbf{Decode and sum outputs}: 
    $
    \hat{\mathbf{y}}^{(l)} \gets \hat{\mathbf{y}}^{(l-1)} + \decoder{l}(\htot{l})
    $
\EndFor
\State \textbf{Return:} $\hat{\mathbf{y}}^{(\bar{l})}$
\end{algorithmic}
\end{algorithm}

Once the multi-fidelity networks are trained offline, they can be used to estimate high-fidelity outputs from the available low-fidelity inputs during the online prediction phase.
The availability of low-fidelity input data depends on the online cost of assimilating or computing those data, given a certain computational budget or a time frame in which the progressive multi-fidelity model needs to operate.
Therefore, if at testing time the budget allows for acquiring low-fidelity inputs $\{\mathbf{x}^{(l)}\}_{l\leq\bar{l}}$ up to level $\bar{l}$, the model will produce the corresponding prediction $\hat{\mathbf{y}}^{(\bar{l})}$.
The online deployment procedure is summarized in Algorithm~\ref{alg:online}.
{In practical time-series applications, some input modalities may not be consistently available at every time step, therefore the number of available inputs $\bar{l}$ can vary over time.
Algorithm~\ref{alg:online} is applied independently at each time step using the currently available modalities, producing a prediction $\hat{\mathbf{y}}^{(\bar{l})}$ that is progressively refined according to the inputs available at the given instant. This approach enables automatic adaptation to missing modalities, efficiently navigates the hierarchy of available data, and ensures continuity of operation.}

The hierarchical ordering here is based on the \textit{availability} of input data rather than on data accuracy, as in traditional multi-level or multi-grid methods, where all data come from a single source and the hierarchy is determined by a control parameter balancing accuracy and computational cost.
In our framework, data may come from multiple sources or modalities, and the hierarchy is defined solely by the availability of the data at the prediction stage \cite{lips2025soft}. {We note that the proposed online deployment assumes a strict hierarchy in data availability, i.e., low-fidelity inputs must be available consecutively from level $0$ up to a maximum level $\bar{l}$.
If data at a given level are missing, higher-level inputs cannot be exploited within the current progressive formulation, as the proposed strategy relies on the sequential concatenation of latent representations.
Extensions aimed at relaxing this hierarchical constraint are briefly discussed in the final remarks in Section~\ref{sect: conclusions}.
}

\subsection{Case studies}
To demonstrate the versatility and robustness of our progressive multi-fidelity framework, we present three case studies spanning both synthetic and real-world scenarios:
\begin{itemize}
    \item \textit{Reaction–diffusion PDE problem}. A parametric, spatio-temporal system where high-fidelity simulations of spiral wave dynamics are reconstructed from coarse, noisy low-fidelity simulations with parametric bias.
    \item \textit{Navier-Stokes PDE problem}. A computational fluid dynamics benchmark leveraging hierarchical low-fidelity inputs (drag and lift coefficients, outlet sensors, and partial-domain snapshots) to reconstruct unsteady flow behavior.
    \item \textit{Air pollution monitoring}. A real-world case using sensor data that combine temperature, humidity, and co-pollutant measurements from low-cost devices to estimate expensive benzene concentrations, despite missing or unreliable low-fidelity signals.
\end{itemize}

These examples address key challenges in scientific machine learning: integrating multi-modal data (full-field simulations, boundary sensors, and scalar time series), mitigating low-fidelity imperfections (noise, model bias, and sparse coverage), and enabling reliable extrapolation beyond training regimes (temporal forecasting and parameter-space generalization). Across all cases, we quantify how each fidelity level improves predictions, balancing accuracy against computational or experimental costs. Uncertainty quantification highlights regimes where low-fidelity inputs are insufficient -- critical for real-world deployment.


\section{Application to simulation data (I): a nonlinear Reaction–Diffusion system}
\label{sec: RD}

In this example, we aim to construct a surrogate model capable of generating spatio-temporal solutions of a nonlinear reaction–diffusion system, governed by the following partial differential equations:
\begin{equation}
    \begin{aligned}
&\dot{u}=\left(1-\left(u^{2}+v^{2}\right)\right) u+\mu\left(u^{2}+v^{2}\right) v+D\left(u_{x x}+u_{y y}\right),\\
&\dot{v}=-\mu\left(u^{2}+v^{2}\right) u+\left(1-\left(u^{2}+v^{2}\right)\right) v+D\left(v_{x x}+v_{y y}\right).
\end{aligned}
\label{eq: RD_eqs}
\end{equation}
The solution $[u,v]^T(x,y,t)$ represents two oscillating modes that generate spiral waves. 
The system \eqref{eq: RD_eqs} is defined over a spatial domain $(x,y)\in [-L,L]^2$ for $L=10$ and a time span $t\in [0,T]$ for $T=80$, where $\mu$ and $D$ are parameters that respectively regulate the reaction and diffusion behaviors of the system.  
We prescribe periodic boundary conditions, and the initial condition is defined as
\begin{equation*}
  \begin{aligned}
     u(x,y,0) = v(x,y,0) = \tanh{\left(\sqrt{x^2 + y^2} \cos{\left((x+iy)-\sqrt{x^2 + y^2}\right)}\right)}\, . 
  \end{aligned}
\end{equation*}

Our goal is to approximate the time evolution of the solution component $u$ as functions of the reaction parameter $\mu\in \mathcal{P} =[0.5, 1.5]$ with a fixed diffusion coefficient $D = 0.05$. 

\subsection{Multi-fidelity setting}
\label{sec: mf-setting RD}
\begin{figure}
    \centering
    \includegraphics[width=1.\linewidth]{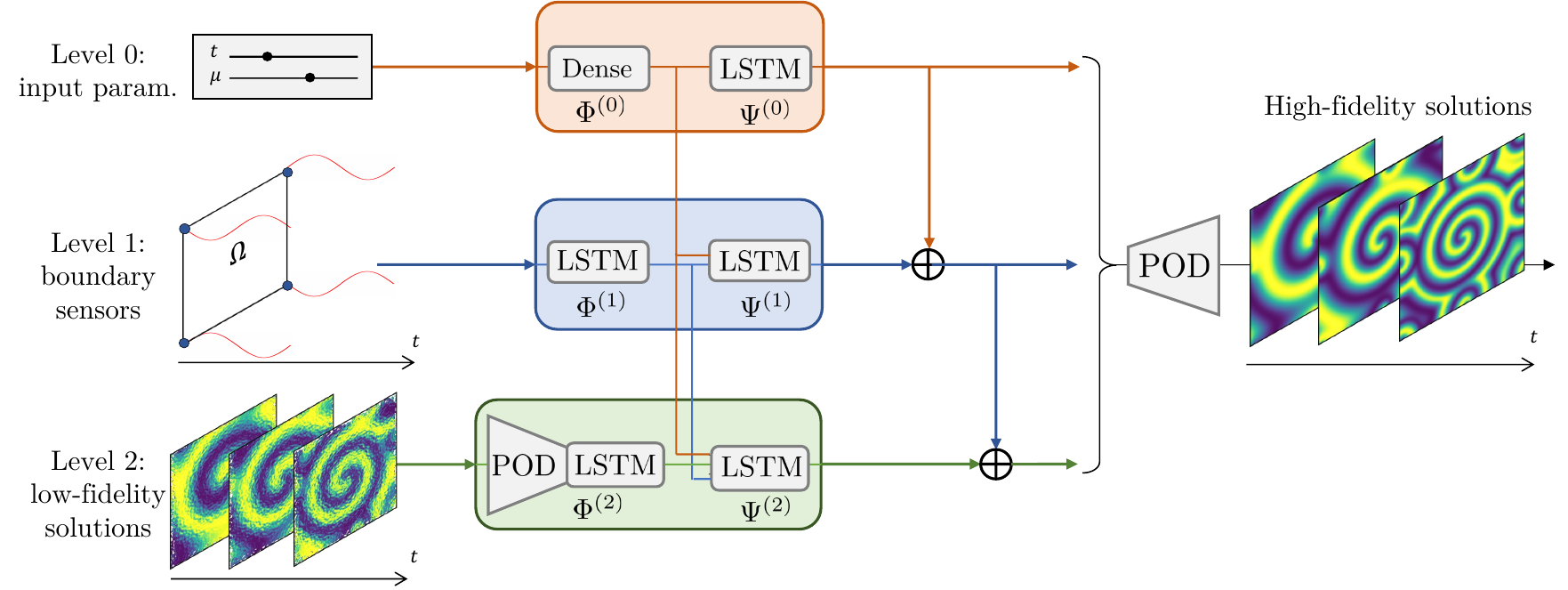}
    \caption{Multi-fidelity model for generating high-fidelity solutions to a reaction-diffusion system. At level 0 a deep-learning surrogate is constructed to estimate spatio-temporal solutions from the control parameters (time and reaction coefficient). In subsequent levels, information from artificial boundary sensors and cheap, low-resolution solutions are progressively
    integrated into the model. Encoder networks $\{\encoder{i}\}_{i=0}^2$ process data from different modalities in a compatible latent representations. The decoder networks $\{\decoder{i}\}_{i=0}^2$ at each level reconstruct spatio-temporal solutions from the latent representations by means of LSTM networks and POD. }
    \label{fig: train_RD}
\end{figure}

We aim to construct a multi-fidelity surrogate to efficiently evaluate high-resolution solution $u$ over the whole spatio-temporal domain, as the reaction parameter $\mu$ varies, exploiting cheaply obtained, low-fidelity output quantities.
In particular, we consider a multi-fidelity surrogate structured in three levels as illustrated in Fig.~\ref{fig: train_RD}. All levels are trained to output the HF solutions, while they differ in the quality and modality of the data provided as input: 

\begin{itemize}
    \item \textbf{Level 0 (input parameters).} A dense FF encoder network $\encoder{0}$ is trained by providing as input data time $t$ and diffusion parameter $\mu$. No information about the solution is given as input. 
    \item \textbf{Level 1 (boundary sensors).} Next level of the surrogate is trained on the temporal evolution of the solution at the domain's vertices. 
    These data artificially simulate information from sensors placed at the domain's boundary. 
    As we need to process time series data, we employ as encoder an LSTM neural network $\encoder{1}$. 
    \item \textbf{Level 2 (LF full-field solutions).} 
    At this level, we exploit the whole LF information, \i.e. we input the LF solutions over the entire spatial domain. 
    As full-field, time series data are given as input, the encoder $\encoder{2}$ needs to extract the dominant spatial patterns as well as encode temporal features. To this end, we combine proper orthogonal decomposition (POD) with LSTM networks. 
    LF data are used to construct a LF POD basis onto which they are projected. The resulting time dependent expansion coefficients are then encoded by an LSTM network.
\end{itemize}

As the output we aim to predict is the spatio-temporal HF solutions, the decoder $\{\decoder{i}\}_{i=0}^2$ combine temporal and spatial encoding, using LSTM networks and POD, respectively -- mirroring the encoder structure of Level 2. 
Thus each level is trained to predict the HF POD coordinates, from which the entire spatial solution can be reconstructed. 
Note that the POD basis employed in the decoders is computed from HF snapshots, thus different from the LF POD basis in $\encoder{2}$. 

In this application, the quality of the input information determines the hierarchical order. 
Note that each level processes a dataset of different modality (vector valued, time-series, spatio-temporal snapshot data) through encoders tailored to the input data fusing their latent representation.
{We note that Level~0 corresponds to a standard single-fidelity surrogate setting, where the HF solution is learned using only HF data and the associated control parameters. The flexibility of the proposed framework allows the encoder and decoder components to be replaced, so that existing deep-learning-based HF surrogates (e.g., \cite{fresca2020comprehensive,hesthaven2018non}) can be embedded as the starting level in the hierarchy.}

\paragraph{Dataset}
Data are computed numerically by solving the PDEs \eqref{eq: RD_eqs} using the Fourier spectral method \cite{trefethen2000spectral} with time step $\Delta t = 0.05$. 
Two distinct datasets are generated for use as input and output to the MF model, respectively, and are referred to as LF and HF based on the following characteristics:

\begin{itemize}
    \item[-] LF solution data are generated on a coarse equispaced spatial grid with $n_\texttt{LF} = 32$ points in each direction, while a fine grid with $n_\texttt{HF} = 100$ is adopted for the HF data. This reduces computational costs and allows LF data to be calculated much faster, at the price of worse accuracy.
    \item[-] LF solutions are evaluated at a corrupted diffusion coefficient $D_\texttt{LF} = 0.1$, instead of $D_\texttt{HF} = D = 0.05$. This represents a bias in the LF modeling in terms of the physical property of viscosity and is employed to emulate model error. 
    \item[-] 
    {LF data are additionally contaminated by multiplicative log-normal noise to simulate observation error. In this work, we consider $\varepsilon \sim \mathrm{LogNormal}(0, 0.8^2),$ i.e., $\varepsilon = \exp(\xi), \xi \sim \mathcal{N}(0, 0.8^2).$
    }
\end{itemize}

To train offline the model, a small amount of expensive HF solution data are computed  over a shorter time horizon $T_\text{train}=40 < T = 80$, at a limited set of parameter locations. 
In particular, $N_\mu = 10$ $\mu$-values are selected over an equispaced grid of $\mathcal{P}$. 
We test the progressive MF method accuracy in predicting the HF solutions over an unseen set of parameters in $\mathcal{P}$ over the whole time window $[0,T]$.
The input dimensions of the data at different fidelity levels are 
$d_\text{in}^{(0)} = 2$, $d_\text{in}^{(1)} = 4$, and $d_\text{in}^{(2)} = 1024$, 
corresponding to the time and parameter values at Level~0, four vertex sensors at Level~1, 
and the low-fidelity spatial degrees of freedom at Level~2. 
The output dimension of the high-fidelity solution is $d_\text{out} = 10000$. 
The dimensionality of both the Level~2 input and the high-fidelity output data is reduced using POD, 
retaining 9 and 11 modes, respectively, capturing about 90\% of the total variance. {The POD bases are constructed from the same snapshot data used for training the surrogate model.}
Uncertainty bounds are computed from the statistical moments of trajectories obtained using ensemble-based methods over $m=30$ retrainings, each with randomly initialized network weights.
More details about the implementation are provided in \ref{sec: appendix}.

\subsection{Results}
\begin{figure}[t!]
    \centering
    \includegraphics[width=1.\linewidth]{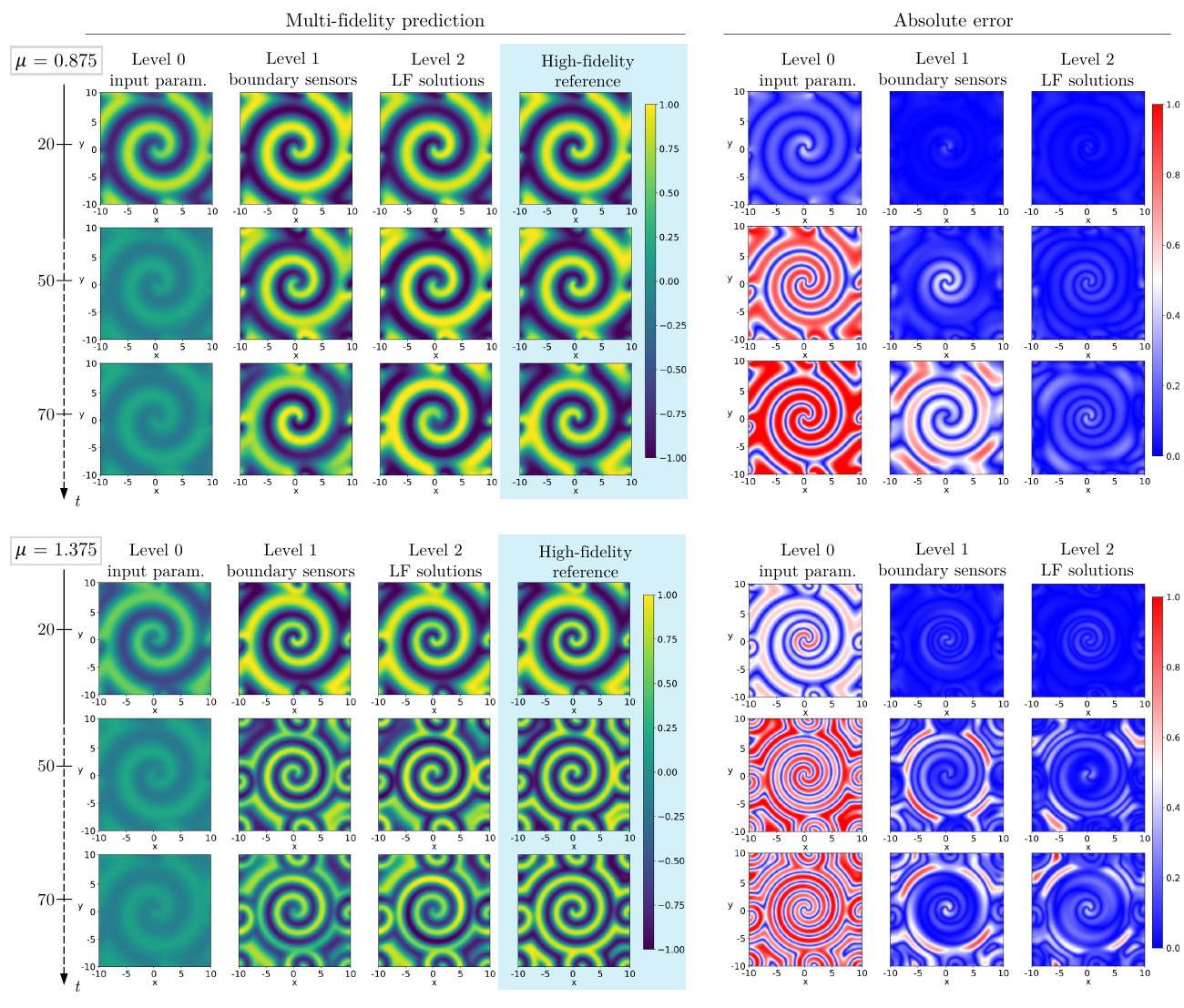}
    \caption{Progressive multi-fidelity prediction (left) for each fidelity level and absolute error (right) with respect to the high-fidelity reference solutions of the reaction-diffusion problem.
    The illustrated snapshots refer to one interpolated time instance $t=20$ and two extrapolated time instances $t\in\{50,70\}$ (being $T_\text{train}=40$ the end of the HF training coverage) for two testing values of reaction parameter $\mu\in\{0.875,1.375\}$ (top and bottom plots, respectively), unseen during the training.}
    \label{fig:RD_pred}
\end{figure}

For unseen testing values of the reaction coefficients $\mu \in \{0.875, 1.375\}$ and for different time instances $t\in\{20,50,70\}$, we show in Fig.~\ref{fig:RD_pred} the reconstructed HF solution fields predicted by the progressive MF model at the different levels, along with the corresponding absolute errors.
We notice a progressive improvement of the performance of the surrogate predictions with corresponding reduction in the error as we move up the hierarchy of the surrogate. 

At level 0, the predictions are generally inaccurate, as the model attempts to reconstruct the entire solution field solely from the control parameters (i.e., time $t$ and reaction coefficient $\mu$). This task is particularly challenging due to the nonlinearity and high dimensionality of the parameter-to-solution map in a supervised regression setting with limited training data. 
As a result, in a single-fidelity context where no additional (LF) information is available, the model fails to extrapolate to unseen temporal scenarios, leading to poor forecasting performance. 
However, for time instances within the temporal window covered during training (albeit for unseen $\mu$), level 0 predictions remain acceptable, capturing the overall solution patterns despite some loss in accuracy (see results for $t=20$ in Fig.~\ref{fig:RD_pred}).


At level~1, temporal evolution data of the solution at the four vertices of the spatial domain are provided as additional inputs. {Although this information is limited to the domain boundaries, it leads to a substantial improvement in prediction accuracy. In this setting, oscillations at the boundary are highly informative of the solution behavior over the entire domain, thereby explaining the marked performance gain obtained when boundary data are incorporated within the multi-fidelity framework.}

Finally, at level 2, the MF framework integrates fast and inexpensive LF solutions -- albeit coarse but defined over the entire spatial domain -- together with the knowledge acquired at previous levels. This enables the model to predict highly accurate solutions of the physical system, showing excellent agreement with the HF reference and substantially reduced prediction errors. 
Accessing this level during the online inference phase requires executing the LF solver to provide the input solutions, and feasibility depends on time or budget constraints. 
In this example, the computational time required to run the full LF simulation is approximately $0.58\text{s}$\footnote{Evaluated on a workstation with an AMD Ryzen 9 5950X 16-core processor.}, after which the progressive MF surrogate reconstructs the HF resolution, yielding an online speed-up of about $37\times$ compared to solving the HF system directly.
Notably, the LF inputs are computed using a corrupted model affected by both parametric error (incorrect diffusion coefficient) and measurement noise (lognormal distribution), demonstrating the robustness of the neural network in mitigating both model and observational biases
Table~\ref{tab: tab_RD} provides a quantitative comparison of relative errors with respect to reference HF test solutions. 

\begin{table}[t]
\caption{ Relative error with respect to the HF reference solution, $\mathbf{y}$, is evaluated for the MF estimates, $\hat{\mathbf{y}}^{(l)}$, at each level $l$, as $\text{err}_{\%}^{(l)}= \frac{100\%}{N_\text{test} }\sum_{n=1}^{N_\text{test} } \frac{ \norm{\mathbf{y}(t_n,\mu_n) - \hat{\mathbf{y}}^{(l)}(t_n,\mu_n) }_{2}}{\norm{{\mathbf{y}}(t_n,\mu_n)}_{2}}$, where $N_\text{test}$ is the number of time-parameter combinations in the test set.
{Additionally, we report the relative error on the predicted POD expansion coefficients, shown in Fig.~\ref{fig: pod_RD}. This metric quantifies the surrogate model error in the reduced space and excludes the contribution of the spatial reconstruction error induced by POD truncation.}
}

\centering
\begin{tabular}{c|c|c|c}
\hline
& Level 0 (input param.) & Level 1 (boundary sensors)  & Level 2 (LF solutions) \\
\hline\hline
Relative error           & 81.3\%      &      29.3\%          &      18.6\%         \\ \hline
{Relative error (POD coeff.)}  & {80.9\%}      &      {25.7\%}          &      {11.8\%}         \\ \hline
\end{tabular}
    \label{tab: tab_RD}
\end{table}

\begin{figure}[t!]
    \centering
    \includegraphics[width=1.\linewidth]{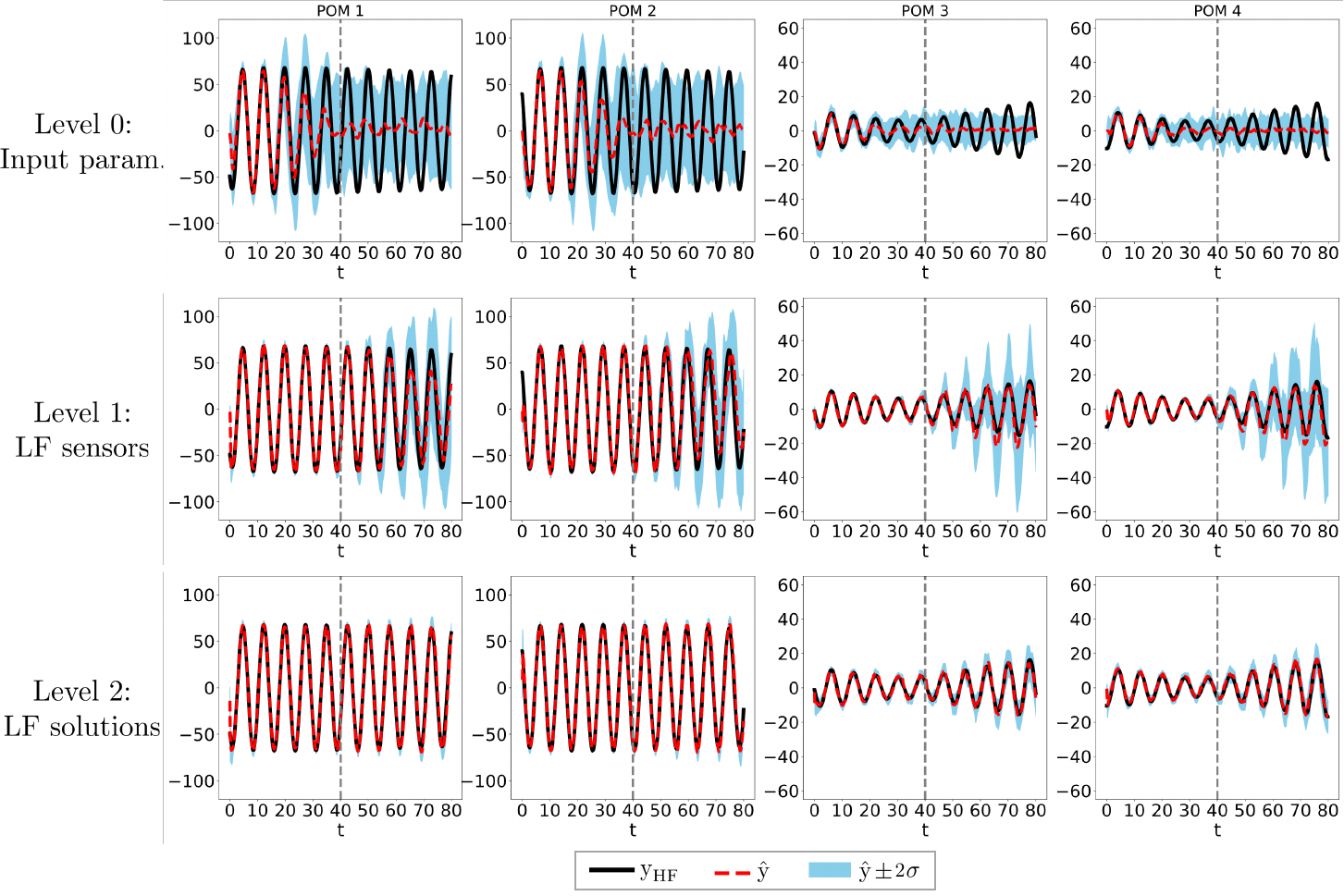}
    \caption{Comparison of the time evolution of the expansion coefficients of the POD modes (POM) in the reaction-diffusion example between the HF reference and the MF predicted solution. For the MF predictions, mean  $\hat{y}$ and two standard deviation $\sigma$ band over the different trainings are shown, providing uncertainty quantification. The vertical dashed line at $T_\text{train}= 40$ indicates the end time of HF data coverage in training, i.e., no HF information is available for $t\in [T_\text{train}, T].$}
    \label{fig: pod_RD}
\end{figure}

As discussed in Sect.~\ref{sec: mf-setting RD}, the decoders at each level estimate the time evolution of the HF POD coefficients, which are then spatially reconstructed by projecting back using the POD spatial basis.
To illustrate the forecasting performance across different levels for unseen parameter values, Fig.~\ref{fig: pod_RD} shows the temporal evolution of several POD coefficients predicted by the LSTM network of decoders.\\
{We observe that, within the training time interval $[0, T_\text{train}]$, with $T_\text{train}=40$, the predictions at the higher fidelity levels closely match the reference solution, indicating that the proposed architecture is effective in interpolation regimes.
At level 0, noticeable deviations from the true response already arise within the training window, reflecting the intrinsic limitations of this level, which relies solely on control parameters and does not incorporate any dynamical or physical information.
As the prediction horizon extends beyond the training interval into $[T_\text{train},T]=[40,80]$, the performance at level 0 further deteriorates, as expected.
This behavior is consistently accompanied by wide uncertainty bounds, which appropriately reflect the increased epistemic uncertainty associated with this lowest-fidelity model.}
In contrast, levels 1 and 2 show a marked improvement in forecasting performance, accurately tracking the system dynamics over long-term horizons for unseen parameters. 
This clearly demonstrates how the integration of physically meaningful LF data helps mitigate the lack of physical consistency often observed in purely data-driven models, thus enabling robust extrapolation capabilities.
As expected, transitioning from level 1 to level 2 results in further gains in predictive accuracy and a noticeable reduction in uncertainty. 
This improvement, however, comes at the cost of running LF solvers over the full spatial domain.

{The uncertainty estimates further enhance interpretation by highlighting regions where predictions are less reliable. Elevated uncertainty consistently aligns with regimes characterized by limited informative inputs (lower-fidelity information) or temporal extrapolation beyond the training window. In particular, at level~0 the uncertainty rapidly increases during long-term forecasting and spans the full oscillatory range of the POD coefficients, reflecting the model’s intrinsic limitations. At higher fidelity levels, uncertainty remains small in interpolation regimes and increases with the prediction horizon, capturing the accumulation of epistemic uncertainty in the recurrent LSTM forecasts. Overall, this behavior shows how the ensemble-based UQ acts as a practical safeguard against overconfident predictions in data-scarce or extrapolative settings.}

{We note that the POD basis used to reduce the dimensionality of the HF outputs imposes an intrinsic upper bound on the achievable accuracy of the reconstructed solution in physical space.
To disentangle the different error contributions, we additionally report in Table~\ref{tab: tab_RD} the relative error on the predicted POD expansion coefficients, which isolates the surrogate modeling error in reduced space and excludes the contribution of POD truncation.
The observed reduction of POD coefficient error across fidelity levels mirrors the improvement in the reconstructed solutions, particularly at Levels~0 and~1, indicating that the dominant source of error in this example arises from the surrogate’s ability to predict the reduced dynamics rather than from POD truncation.
At Level~2, the discrepancy between the coefficient error and the reconstructed solution error becomes non-negligible, reflecting the contribution of POD truncation.
This effect could be mitigated by retaining a larger number of POD modes; however, this would require predicting a higher-dimensional output, typically associated with lower-variance modes exhibiting more complex temporal behavior, thereby increasing the difficulty of the regression task.
In practice, we observed that increasing the number of retained modes from 11 to 16 led to a higher overall reconstruction error at level~2 (26.2\% versus 18.6\%), indicating that the chosen truncation represents a well balanced trade-off between expressiveness and learnability.
}

\section{Application to simulation data (II): Navier-Stokes fluid flows}
\label{sec: NS}

We consider a two-dimensional fluid flow around a cylinder --- a benchmark problem in computational fluid dynamics. For a viscous, incompressible Newtonian fluid flow, the problem is governed by the following Navier-Stokes equations
\begin{equation}
    \begin{aligned}
           \rho \frac{\partial \textbf{v}}{\partial t} + \rho \textbf{v}\cdot \nabla\textbf{v} - \nabla \cdot \bm{\sigma}(\textbf{v},p) &=  \textbf{0}, \qquad &&(\bm{x},t)\in \Omega \times (0,T) \, ,  \\
    \nabla \cdot \textbf{v} &= 0, \qquad &&(\bm{x},t)\in \Omega \times (0,T) \, ,
    \end{aligned}
    \label{eq: NS}
\end{equation}
where $\textbf{v}(\bm{x},t)$ and $p(\bm{x},t)$ represent the velocity and pressure field, respectively, and $\rho = 1.0 \text{ kg/m}^3$ is the fluid density, and $\bm{\sigma}(\vb{v},p)=-p\vb{I}+2\nu\bm{\epsilon}(\vb{v})$ is the stress tensor, where $\bm{\epsilon}(\vb{v})$ denotes the strain tensor.
The kinematic viscosity is defined as $\nu = 1 / Re$, where the Reynolds number $Re$ is the system parameter of interest.
Here, the spatial domain $\Omega = (0, 2.2) \times (0, 0.41) \text{\textbackslash} B_r(0.2,0.2)$ ($r = 0.05$) represents a 2D channel with a cylindrical obstacle, where no-slip boundary conditions are prescribed on the side walls and on the cylinder, a parabolic inflow on the inlet, and an open boundary condition at the outlet. As the initial condition, we consider the fluid at rest.

In the present study, we consider $T = 20\text{s}$ and $\mu = Re \in \mathcal{P} = [30,100]$. When $Re < 49$, the flow exhibits a {\textit{steady}} behavior; for larger Reynolds numbers, the flow transitions to an \textit{unsteady} state, and a pair of vortices forms in the wake of the cylinder, oscillating periodically between top and bottom sides \cite{RAJANI20091228, zdravkovich1997flow}.

\subsection{Multi-fidelity setting}
\label{sec: mf-setting NS}
Our goal is to estimate the spatio-temporal evolution of the velocity magnitude $v(\bm{x},t)=\norm{\textbf{v}(\bm{x},t)}_2$  as a function of the Reynolds number, from the following low-fidelity information:

\begin{itemize}
    \item \textbf{Level 0 (input parameters).} A FF encoder network $\encoder{0}$ is trained by providing as input data time $t$ and Reynolds number $Re$. 
    \item \textbf{Level 1 (drag and lift).} The next level is trained on the drag and lift time series, two dimensionless, time-dependent coefficients that quantify forces acting on the obstacle \cite{schafer1996benchmark}.
    Although they are informative on the system’s temporal evolution, they are output quantities that do not provide a direct measure of the solution field. 
    We employ an LSTM neural network $\encoder{1}$ as encoder. 
    \item \textbf{Level 2 (outflow).} Solution values at the outflow are provided as input, simulating sensors on the outlet boundary.
    The evolution of the outflow front measured at equispaced spatial points on the outlet is processed as a vector of time-series by an additional LSTM encoder $\encoder{2}$.
    \item \textbf{Level 3 (partial domain).} This level simulates a video sensor that assimilates snapshots of the velocity magnitude over a partial domain surrounding the obstacle. 
    Here, $\encoder{3}$ employs an autoencoder architecture with LSTM layers to encode the high-dimensional spatio-temporal data.
\end{itemize}

\begin{figure}
    \centering
    \includegraphics[width=1.\linewidth]{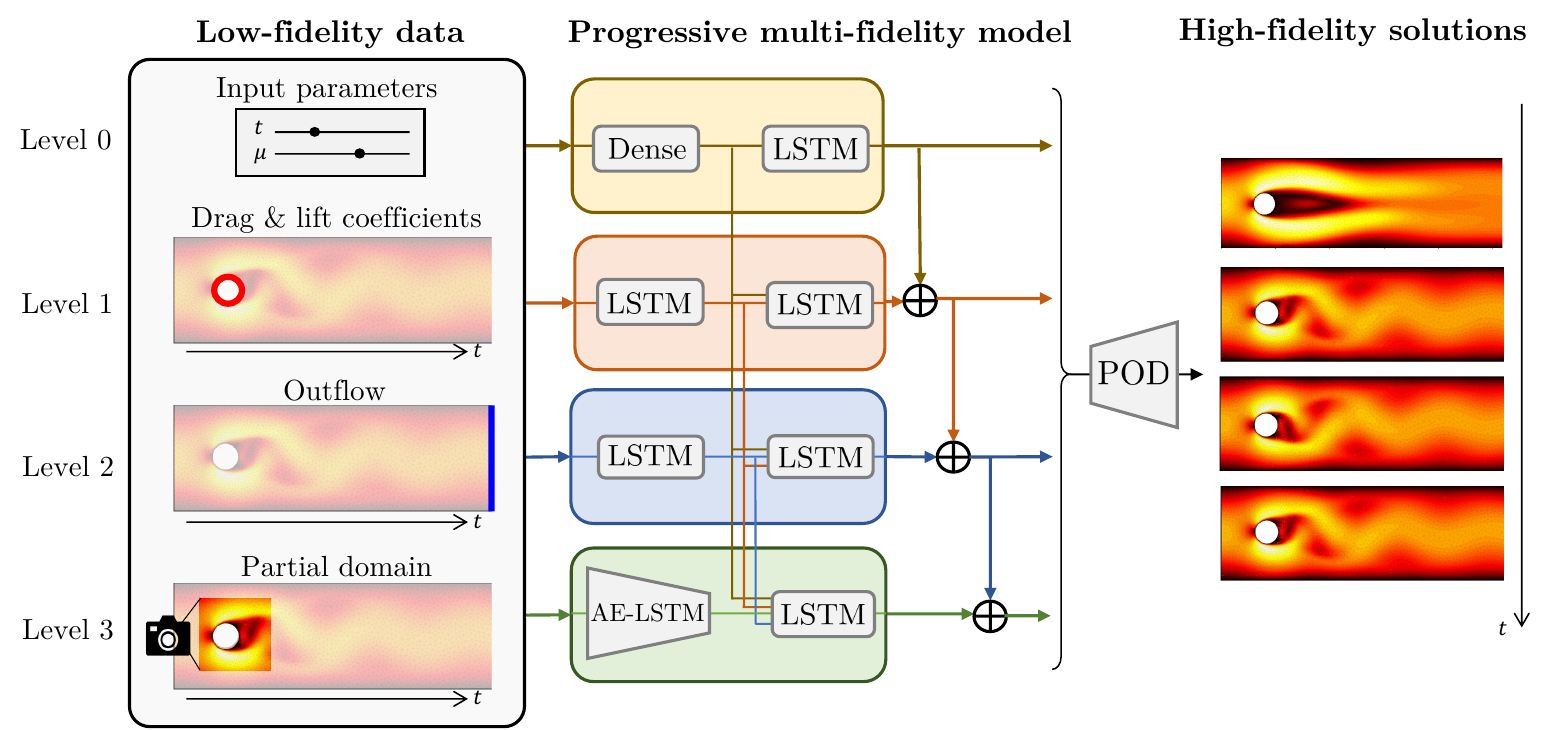}
    \caption{Multi-fidelity model for generating high-fidelity solutions to the Navier-Stokes equations. At level 0 a deep-learning surrogate is constructed to estimate spatio-temporal solutions from the input parameters (time and Reynolds number coefficient). At level 1 information about the time evolution of drag and lift coefficients, expressing the forces acting on the obstacle wall, are integrated. At level 2 information about solution outflow, representing boundary sensors, is included. Finally, spatio-temporal snapshots of a limited portion of the domain (surrounding the obstacle) are assimilated by the model at level 3. Encoder networks $\{\encoder{i}\}_{i=0}^3$ process data from different modalities into compatible latent representations. The decoder networks $\{\decoder{i}\}_{i=0}^3$ at each level reconstruct spatio-temporal solutions from the latent representation by means of LSTM networks and POD.}
    \label{fig:train_NS}
\end{figure}

To keep the training computationally light, we consider, as in the previous application, a combination of POD and LSTM networks as $\{\decoder{i}\}_{i=0}^3$ to decode the spatio-temporal features of the high-fidelity velocity magnitude. A representation of the multi-fidelity architecture is shown in Fig.~\ref{fig:train_NS}.

\paragraph{Dataset}
Low-fidelity input and high-fidelity output data are generated through a finite element approximation of \eqref{eq: NS} with quadratic finite elements and backward differentiation formula provided by the MATLAB library \texttt{redbKIT} \cite{negri2016redbkit}. 
We consider a training set computed over a short time window $T_\text{train} = 15\text{s} <T = 20 \text{s}$ at $N_\mu = 15$ Reynolds numbers equispaced over $\mathcal{P}$.
High-fidelity data are computed over a mesh with 16478 triangular elements and $d_\text{out}=8239$ vertices, while the temporal discretization is with step size $\Delta t = 5 \text{ms}$. 
The high-fidelity output data consists of the time evolution of the velocity magnitude at the $d_\text{out}$ spatial degrees of freedom (dof). {The dimensionality of the high-fidelity output data is reduced using POD, retaining 32 modes, which capture approximately 97\% of the total variance.}
The input dimensions of low-fidelity data are $d_\text{in}^{(0)}=2, d_\text{in}^{(1)}=2$ (time $t$ and parameter $\mu$ at level 0, and drag and lift at level 1), and $d_\text{in}^{(2)}=20, d_\text{in}^{(3)}=577$ (spatial degrees of freedom on the outlet boundary and on a portion of the domain surrounding the obstacle; see Fig.~\ref{fig:train_NS}).
Further implementation details and training specifications are provided in \ref{sec: appendix}.

\subsection{Results}
\begin{figure}[t!]
    \centering
    \includegraphics[width=1.\linewidth]{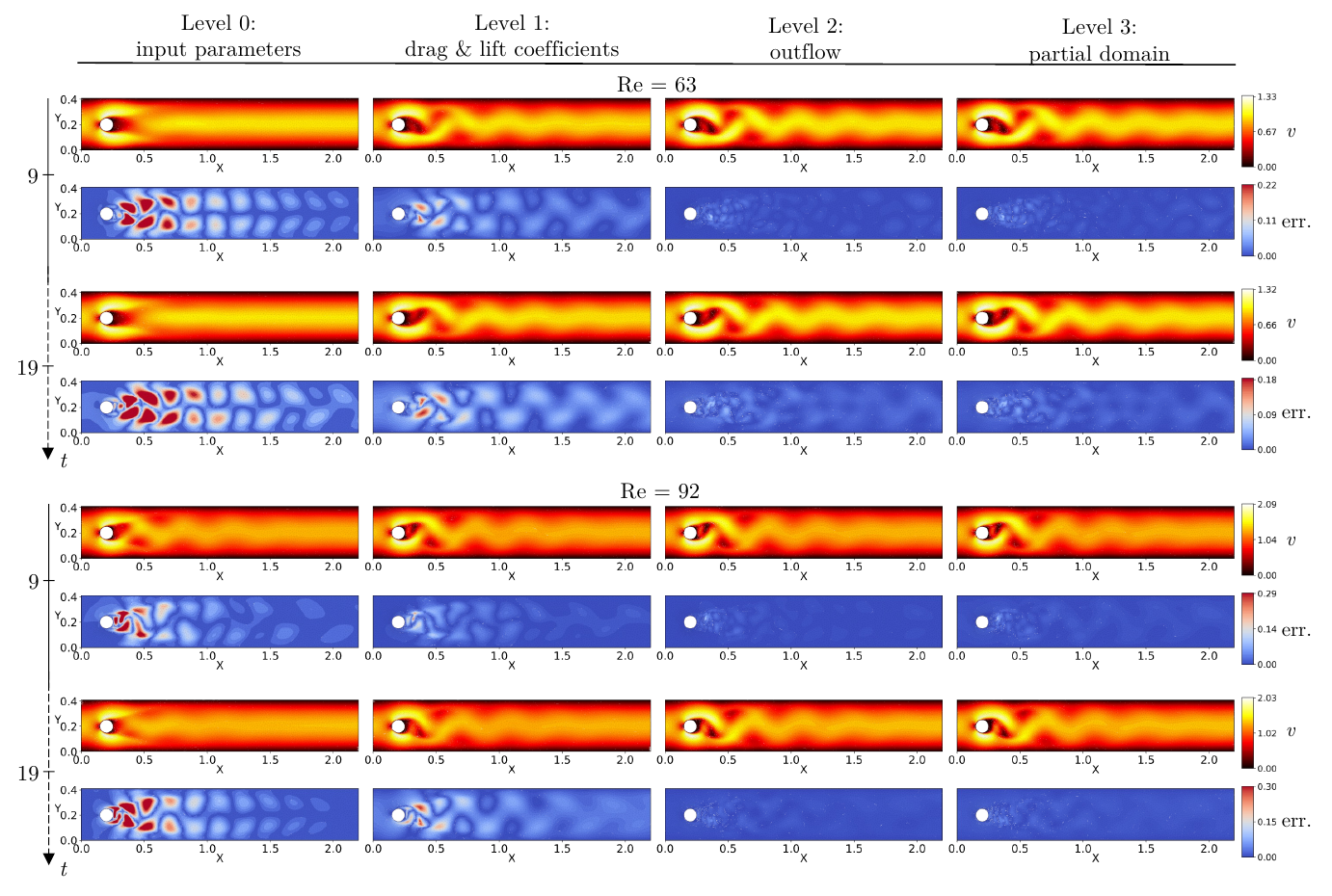}
    \caption{Progressive multi-fidelity prediction of velocity magnitude $v$ for each fidelity level and absolute error with respect to the high-fidelity reference solutions of the Navier-Stokes problem.
    The illustrated snapshots refer to one interpolated time instance $t=9\text{s}$ and one extrapolated time instance $t=19\text{s}$ (with $T_\text{train}=15\text{s}$ marking the end of the HF training coverage) for two testing values of Reynolds number $\mu\in\{63,92\}$ (top and bottom plots, respectively), unseen during the training.}
    \label{fig:results_NS}
\end{figure}

For unseen Reynolds numbers $Re\in\{63,92\}$ and different time instances $t\in\{9\text{s},19\text{s}\}$ (with $[0\text{s},15\text{s}]$ the training window), Fig.~\ref{fig:results_NS} shows the predicted high-fidelity velocity magnitude at each level of the progressive MF model, along with the corresponding absolute errors.
At level 0, the surrogate relies solely on $(t,Re)$ as inputs and can only reproduce the average stationary behavior of the flow, missing the unsteady vortex-shedding dynamics.
Incorporating drag and lift coefficients at level 1 significantly improves predictions, despite the absence of any direct measurements of the solution field.
Nonetheless, non-negligible errors remain in the wake region, where nonlinear effects dominate.
At level 2, additional measurements from sensors on the outlet boundary enable accurate and spatially homogeneous predictions, even beyond the training time window. Notably, the accuracy at $t=19\text{s}$ is comparable to that at $t=9\text{s}$, demonstrating the model’s capability to reliably forecast forward in time..
Remarkably, improvements are observed throughout the domain -- including the cylinder wake, where no sensors are placed -- highlighting the model’s ability to fuse and effectively exploit localized information.
Finally, level 3, which assimilates partial-domain snapshots around the obstacle, yields only marginal gains over level 2, indicating that the outlet boundary data, combined with drag and lift measurements, already provide sufficiently rich information for the neural architecture.
Table~\ref{tab: tab_NS} reports the overall error over the full parametric, spatio-temporal test set, further supporting these conclusions. 
{Additionally, as in the reaction–diffusion example, we report the relative error on the POD expansion coefficients. The small discrepancy between the reduced-space error and the reconstructed full-field error indicates that POD truncation is negligible in this case. Consequently, the overall error is primarily driven by the progressive neural network’s ability to predict the POD coefficients, confirming that the method’s performance is governed by the surrogate model’s accuracy.}

\begin{table}[b!]
\caption{Relative error with respect to the HF reference solution, $\mathbf{y}$, is evaluated for the MF estimates, $\hat{\mathbf{y}}^{(l)}$, at each level $l$, as  $\text{err}_{\%}^{(l)}= \frac{100\%}{N_\text{test} }\sum_{n=1}^{N_\text{test} } \frac{
\norm{\mathbf{y}(t_n,\mu_n) - \hat{\mathbf{y}}^{(l)}(t_n,\mu_n) }_{2}}{\norm{{\mathbf{y}}(t_n,\mu_n)}_{2}}$, where $N_\text{test}$ is the number of time-parameter combinations in the test set.
{Additionally, the table reports the relative error on the predicted POD expansion coefficients.}}
\centering
\begin{tabular}{c|c|c|c|c|c}
\hline
& Level 0 (param.) & Level 1 (drag/lift)  & Level 2 (outflow) & Level 3 (part. domain)\\
\hline\hline
Relative error         & 12.3\%      &      8.73\%          &      6.82\%       & 6.50\%  \\ \hline
{Rel. err. (POD coeff.)}  & {12.3\%}      &      {8.69\%}          &      {6.76\%} & {6.44\%}  \\ \hline
\end{tabular}
    \label{tab: tab_NS}
\end{table}

\section{Application to real measurement data: air pollution forecasting}
\label{sec: air_pollution}

Air pollution poses a critical threat to public health, requiring accurate monitoring of pollutant concentrations in urban areas \cite{mage1996urban}.
Currently, air pollution monitoring is performed by fixed monitoring stations (based on industrial spectrometers), which can provide high-fidelity data with precise measurements but are expensive, bulky, and limited in spatial coverage, making evaluation in complex urban environments challenging.
Low-cost multi-sensor devices, or ``electronic noses'', offer a complementary solution by increasing network density at a lower cost.
However, their accuracy is restricted by the instability and limited selectivity of the solid-state sensors they rely on \cite{de2008field}.

Our goal is to develop a progressive multi-fidelity surrogate model to perform HF estimates of benzene, $\text{C}_6\text{H}_6$, concentration (in $\unit{\micro\gram}/\text{m}^3$) -- a key pollutant related to respiratory illness -- from a hierarchical set of four measurements provided by an LF multi-sensor. 
This example serves as a proof of concept for the development of an MF software capable of recovering HF estimates by calibrating LF measurements, thereby broadening accurate monitoring capabilities in resource-constrained settings.
The LF multi-sensor device \cite{de2008field} is equipped with metal oxide chemoresistive sensors for pollutant measurements, designed to be independently replaceable for ease of maintenance. 
In addition, it includes low-cost commercial sensors for temperature and relative humidity. 
The LF measurements are organized into a multi-fidelity hierarchy. 

\subsection{Multi-fidelity setting}
\label{sec: mf-setting air}

\begin{figure}[t!]
    \centering
    \includegraphics[width=1.\linewidth]{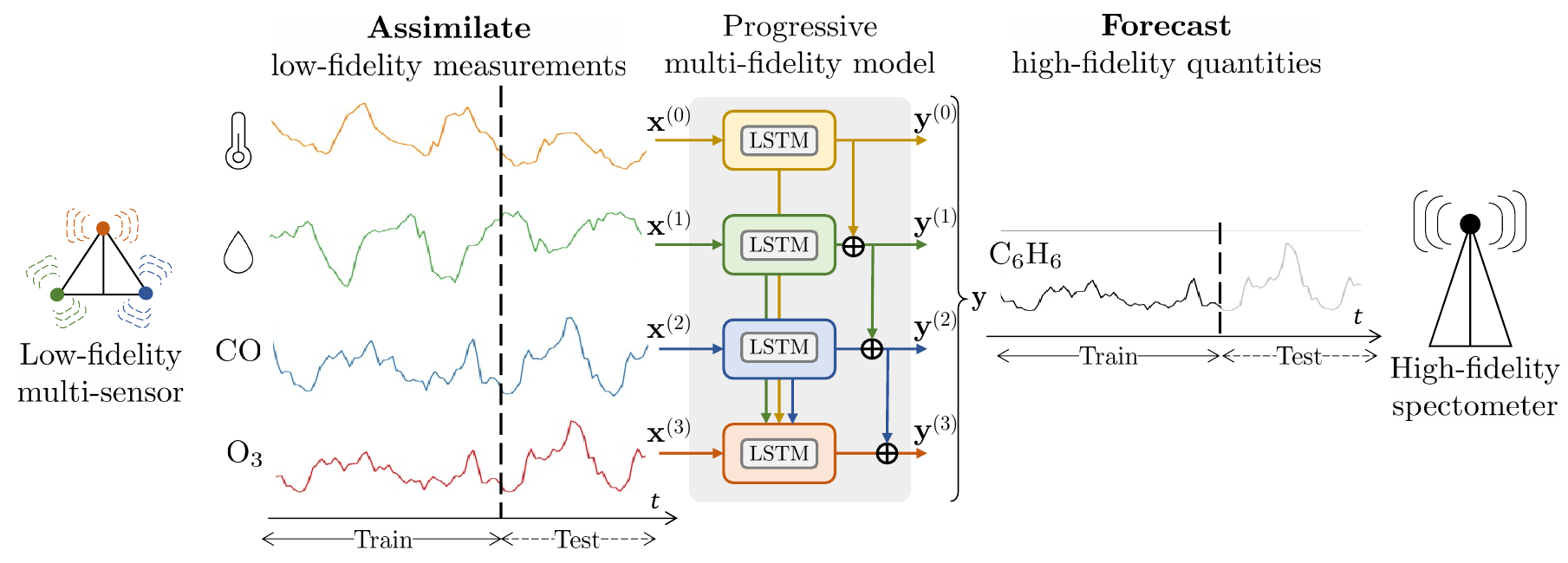}
    \caption{Multi-fidelity set-up for estimating air pollution. A low-fidelity multi-sensor device provides cheap measurements of temperature, humidity and concentrations of $\text{CO}$ and $\text{O}_\text{3}$. A progressive multi-fidelity model is trained to provide a hierarchy of approximations of the output quantity of interest, which is the high-fidelity estimation of the benzene ($\text{C}_\text{6}\text{H}_\text{6}$) measured by an expensive spectrometer. High-fidelity output data are available for a limited time window for training, while outputs are forecast by the multi-fidelity model at different levels of refinement, according to the availability of low-fidelity measurements.}
    \label{fig: airpollution_setup}
\end{figure}

The first two signals, $x^{(0)}$ and $x^{(1)}$, are temperature (in $\text{C}^\circ$) and relative humidity (in $\%$), respectively.
These LF signals are inexpensive to obtain and could capture general trends of heating and meteorological influences, but they are poorly correlated with specific pollutant levels. 
The following two signals, $x^{(2)}$ and $x^{(3)}$, are LF measurements of the concentration 
(in $\unit{\micro\gram} / \text{m}^3$) of related pollutants, CO and $\text{O}_\text{3}$, which do not directly estimate benzene, yet offer valuable complementary information.
Importantly, these LF measurements can be acquired at much lower cost than those from HF industrial spectrometers.
As shown in Fig.~\ref{fig: airpollution_setup}, the LF signals $\{x^{(l)}\}_{l=0}^3$ are provided as input to the progressive multi-fidelity model to estimate the HF benzene concentration $y$ (in $\unit{\micro\gram} / \text{m}^3$).
Outputs at each level are indicated as $\{\hat{y}^{(l)}\}_{l=0}^3$ and are expected to progressively improve in accuracy and reliability as more LF information is assimilated while proceeding up the hierarchy.

\paragraph{Dataset}
Both LF inputs (at each level) and HF outputs are one-dimensional time series measurements.
Therefore all encoder and decoder networks, $\encoder{l}$ and $ \decoder{l}$ respectively, are LSTM networks with dimensions $d_\text{in}^{(l)} = d_\text{h}^{(l)} = d_\text{out}^{(l)} = 1$, for $l=0,\ldots,3$.

Input-output data are available to train the progressive MF model from $T_0 = 2004/10/03$ (YYYY/MM/DD) to $T_\text{train}=2005/01/16$.
All measurements are recorded on an hourly basis.
Each level is trained $m=15$ times to construct an ensemble of models, which are used to provide uncertainty quantification.
After $T_\text{train}$, we assume to assimilate only LF data up to $T_\text{test} = 2005/04/04$ and we test the performance of our model in estimating the HF output in $[T_\text{train}, T_\text{test}].$

{The data used in this example are extracted from the work by De Vito \textit{et al.}~\cite{de2008field} and are publicly available through the \href{https://archive.ics.uci.edu/dataset/360/air+quality}{UCI Machine Learning Repository}. 
LF inputs include chemoresistive gas sensors targeting $\text{CO}$ and $\text{O}_3$, measured by the Pirelli Labs multi-sensor device, designed for continuous urban monitoring with a sensor array, data processing unit, and communication module. 
Two additional slots host commercial temperature and humidity sensors (levels 0 and 1, respectively), which are less prone to drift or replacement, providing robust and continuously available LF signals. 
$\text{CO}$ and $\text{O}_3$ sensors are more sensitive to environmental conditions and aging, and may require replacement or recalibration, constituting higher-level LF inputs (levels 2 and 3) whose availability can vary over time. 
HF outputs correspond to benzene concentrations measured by reference-grade analyzers~\cite{de2008field}. 
All measurements are preprocessed by removing invalid entries (sentinel value $-200$) and filling missing values via linear interpolation independently for each variable. 
Finally, all signals are rescaled to the unit interval. No additional filtering or denoising is applied, ensuring that model evaluation reflects realistic measurement noise and sensor irregularities.}

\begin{figure}[b!]
    \centering
    \includegraphics[width=1.\linewidth]{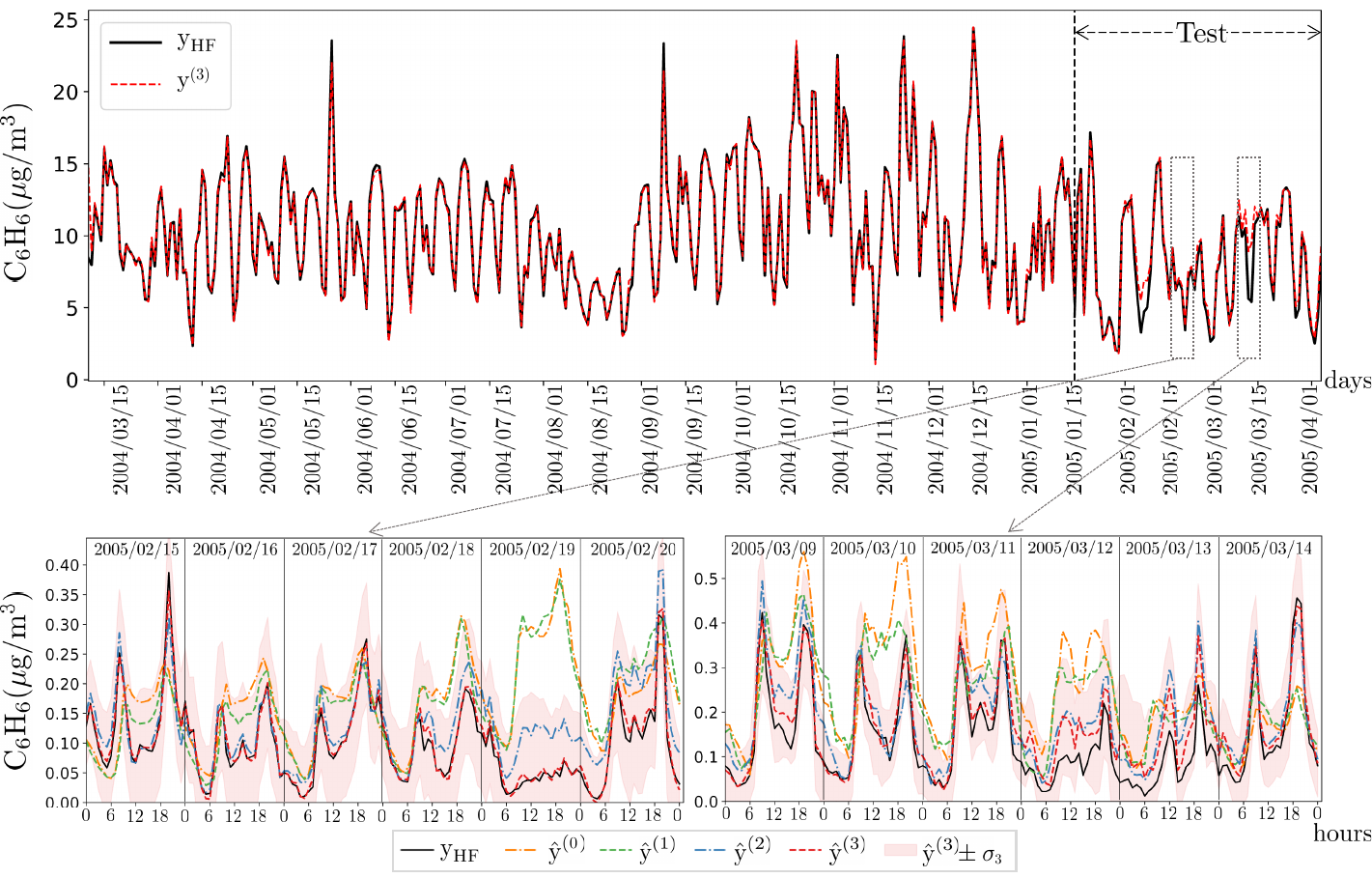}
    \caption{Results for the air pollution test case.
    The top plot shows the comparison between the actual benzene concentration $\text{y}_\text{HF}$ (black line) and the prediction of the multi-fidelity surrogate at the final level $\hat{\text{y}}^{(3)}$ (red dashed line), aggregated daily over the entire time window.  The dashed black line at $T_\text{train} = 2005/01/16$  indicates the end time of HF data coverage in training, i.e. no HF information is available over $t\in[T_\text{train},T_\text{test}].$
    The bottom plot provides two zoomed-in views, highlighting the forecasting capabilities of the multi-fidelity surrogates at hourly frequency across all hierarchy levels.
     The shaded region indicates $\pm$ one standard deviation $\sigma_3$ around the prediction at the final level, offering uncertainty quantification for the multi-fidelity estimates.}
    \label{fig: airpollution_results}
\end{figure}

\begin{table}[t]
\centering
\begin{tabular}{c|c|c|c|c}
\hline
& Level 0 (Temperature) & Level 1 (Humidity)  & Level 2 ($\text{CO}$) & Level 3 ($\text{O}_3$)\\
\hline\hline
Relative error           &  72\%      &      66\%          &      29\%       & 13\%  \\ \hline
\end{tabular}
\label{tab: tab_airpollution}
\caption{Relative error with respect to the HF data, $y$, is evaluated for the MF estimates, $\hat{{y}}^{(l)}$, at each level $l$, as  $\text{err}_{\%}^{(l)}=\frac{100\%}{N_\text{test} }\sum_{n=1}^{N_\text{test} } \frac{
        \norm{{y}(t_n) - \hat{{y}}^{(l)}(t_n) }_{2}}{\norm{{{y}}(t_n)}_{2}}$, where $N_\text{test}$ is the number of time instants in the test set.}
\end{table}

\subsection{Results}
The top plot in Fig.~\ref{fig: airpollution_results} showcases the comparison between the last level prediction of the MF surrogate model and  the reference solution, demonstrating high accuracy in reconstructing HF predictions solely from LF measurements. Data are aggregated on a daily basis.\\
In the bottom plot, two zoomed-in views provide a more detailed analysis of the method's performance across the four levels and at an hourly scale. 
The left subplot focuses on a period where the model performs particularly well (days $2005/02/15-20$), while the one on the right highlights the most significant discrepancies (days $2005/03/9-14$). 
In both scenarios, we observe the progressive enhancement in accuracy as we traverse the hierarchy of the MF model. Each level progressively refines the estimated benzene concentration by integrating the information from the corresponding inputs and all preceding levels.
\\In particular, at the first levels $l=0$ and $l=1$, the model relies on temperature and humidity signals, which offer general environmental context but exhibit limited direct correlation with benzene concentration. 
Despite these limitations, predictions at these levels effectively capture long-term trends and provide reasonably reliable estimates. 
However, discrepancies arise in certain periods, where the lack of pollutant-specific data leads to noticeable deviations from the actual HF reference values. 
Importantly, since temperature and humidity measurements are easy to obtain and consistently available, the progressive MF model ensures continuity in operation. Indeed, even if pollutant sensors ($x^{(2)}$ for CO and $x^{(3)}$ for $\text{O}_3$) require maintenance or break down, the lower-fidelity levels can still provide reasonable estimates, preventing complete model downtime.
\\A substantial improvement is observed at level $l=2$, where LF measurements of CO concentration are incorporated. This additional input introduces crucial chemical context, significantly enhancing the model's ability to resolve short-term fluctuations in benzene concentration. As a result, level 2 predictions align more closely with the HF ground truth, substantially reducing errors observed at previous levels.
\\At the final level, $l=3$, where $\text{O}_3$ concentration is assimilated, the model attains its highest accuracy. The progressive assimilation of information allows this level to make precise corrections, achieving great alignment with HF measurements.
\\Crucially, even in the most challenging time window (days $2005/03/9-14$), the true HF solution is consistently embedded within the model’s uncertainty bounds, highlighting the reliability of our approach and its robustness in real-world scenarios where HF data are not available.
\\Overall, the results validate the effectiveness of our progressive multi-fidelity strategy. 
The hierarchical structure enables the model to make reasonable predictions even at lower-fidelity levels while progressively refining estimates as richer information becomes available.
This adaptability makes our approach particularly suitable for deployment in real-time monitoring systems, where HF data may be limited due to cost or logistical constraints, and ensures operational continuity even in the presence of temporary sensor malfunctions.
\section{Conclusions}
\label{sect: conclusions}
In this work, we introduced a progressive multi-fidelity learning framework designed to construct surrogate models for physical systems by sequentially integrating heterogeneous, multi-modal datasets. 
The core of the approach is a hierarchical architecture that processes data at each fidelity level through tailored encoders, fuses the resulting latent representations, and applies additive corrections to predictions from preceding levels.
This design ensures that knowledge is retained and refined as new data becomes available, effectively mitigating catastrophic forgetting.

We demonstrated the flexibility and effectiveness of our method across three distinct applications: two involving simulated data (a nonlinear reaction-diffusion system and a Navier-Stokes fluid flow problem) and one real-world task (air pollution monitoring).
The results consistently demonstrated that the progressive fusion of information allows the model to efficiently leverage knowledge from low-fidelity data, leading to a systematic increase in prediction accuracy and a corresponding reduction in predictive uncertainty at each level.
Crucially, the framework maintains operational continuity, providing reasonable estimates even when only lower-fidelity inputs are available due to computational, budget, or data assimilation constraints. The integrated uncertainty quantification further enhances the model's reliability, even in extrapolative regimes.
{Regarding generalization, the air pollution application showcases strong capabilities on pure temporal forecasting, while the PDE benchmarks assess spatio-temporal generalization through temporal extrapolation and parameter interpolation. For nonlinear systems, extrapolation across parameter regimes with qualitatively different dynamics may be ill-posed for data-driven surrogates. By leveraging the progressive structure and physically meaningful low-fidelity data, the framework improves generalization in time and across parameter regimes where correlations between fidelity levels remain consistent. Conversely, for parameter regimes with fundamentally different dynamics, generalization gains may be limited, highlighting an important avenue for future study.}\\
Moreover, future work will focus on enhancing the framework's flexibility and computational efficiency. 
{A primary direction is the development of multi-fidelity architectures that do not rely on a strict hierarchy of inputs. 
\paolo{In the current progressive formulation, if an intermediate fidelity level is missing, higher-level inputs cannot be exploited, which can be limiting in multi-modal applications where a clear fidelity ordering cannot be established. The hierarchical assumption also increases computational cost: fidelity levels are added sequentially, so each additional level requires training an additional surrogate level.}
Early-fusion strategies combined with attention mechanisms, for instance, allow the model to dynamically weight the contribution of each available fidelity level, enabling inference with arbitrary combinations of missing or available inputs \cite{jaegle2021perceiver, jaegle2021perceiverio, girdhar2023imagebind}.
However, relaxing the hierarchical structure generally removes the structural guarantees against catastrophic forgetting provided by the progressive training strategy adopted here, where previously trained components are frozen and higher-fidelity information is incorporated exclusively through additive corrections. Designing flexible fusion mechanisms that retain such guarantees remains an open research challenge.}
Another promising direction is to improve efficiency, such that the computationally expensive ensemble-based uncertainty quantification could be replaced with inherent probabilistic architectures, such as Bayesian neural networks or conditional neural processes \cite{garnelo2018conditional, wu2022multi}.

Finally, the generality and effectiveness of the paradigm suggest its applicability to a broader range of scientific computing problems, including inverse problems where the multi-fidelity structure could be naturally embedded within multi-level inference schemes.


\section*{Acknowledgment}
PC has been supported under the JRC STEAM STM-Politecnico di Milano agreement and by the PRIN 2022 Project “Numerical approximation of uncertainty quantification problems for PDEs by multi-fidelity methods (UQ-FLY)” (No. 202222PACR), funded by the European Union - NextGenerationEU. PC also thanks Dr. Johannes Lips (University of Stuttgart) for fruitful discussions on the air pollution example.

AM acknowledges the project “Dipartimento di Eccellenza” 2023-2027 funded by MUR, the project FAIR (Future Artificial Intelligence Research), funded by the NextGenerationEU program within the PNRR-PE-AI scheme (M4C2, Investment 1.3, Line on Artificial Intelligence) and the Project “Reduced Order Modeling and Deep Learning for the real- time approximation of PDEs (DREAM)” (Starting Grant No. FIS00003154), funded by the Italian Science Fund (FIS) - Ministero dell'Università e della Ricerca. AM is member of the Gruppo Nazionale Calcolo Scientifico-Istituto Nazionale di Alta Matematica (GNCS-INdAM).

AF acknowledges the PRIN 2022 Project “DIMIN- DIgital twins of nonlinear MIcrostructures with iNnovative model-order-reduction strategies” 
(No. 2022XATLT2) funded by the European Union - NextGenerationEU.

\appendix
\section{}
\label{sec: appendix}
\setcounter{table}{0}

\begin{table}
\centering
\begin{tabular}{l|ccc}
\toprule
\textbf{Hyperparameter} & \textbf{Reaction--Diffusion} & \textbf{Navier--Stokes} & \textbf{Air--Pollution} \\
\midrule
\textbf{Encoder type} \\
  \quad Level 0 &  Dense & Dense &  LSTM \\
  \quad Level 1 & LSTM &  LSTM & LSTM \\
  \quad Level 2 & POD+LSTM & LSTM &  LSTM \\ 
  \quad Level 3 &  &  AE-LSTM & LSTM\\
\midrule
\textbf{Input dimensions $d_\text{in}$}\\
  \quad Level 0: $d_\text{in}^{(0)}$ & 2 & 2 & 1 \\
  \quad Level 1: $d_\text{in}^{(1)}$ & 4 & 2 & 1 \\
  \quad Level 2: $d_\text{in}^{(2)}$ & 1024 ($N_\text{POD}=9$) & 20 & 1\\ 
  \quad Level 3: $d_\text{in}^{(3)}$ & & 577 & 1 \\
\midrule
\textbf{Latent dimensions $d_h$}\\
 \quad Level 0: $d_h^{(0)}$ & 2 & 2 & 1 \\
  \quad Level 1: $d_h^{(1)}$  & 2 & 2 & 1 \\
  \quad Level 2: $d_h^{(2)}$ & 4 & 2 & 1\\ 
  \quad Level 3: $d_h^{(3)}$ & & 2 & 1\\
\midrule
\textbf{Encoder NN layers/nodes} \\

\quad Level 0  & [31] & [20] & [20,20] \\ 
\quad Level 1  & [31] & [20] & [20,20]\\ 
\quad Level 2  & [31] & [20] & [20,20]\\ 
\quad Level 3  & & [20] & [20,20]\\
\midrule
\textbf{Output dimensions} $d_\text{out}$ \\
\quad same for all levels & 10000 ($N_\text{POD}=11$) & 8239 ($N_\text{POD}=32$) & 1 \\
\midrule
\textbf{Decoder type} \\
\quad same for all levels & LSTM+POD & LSTM+POD & LSTM \\
\midrule
\textbf{Decoder NN layers/nodes} \\
\quad same for all levels & [25, 25, 25] & [20, 20] & [20,20] \\
\midrule
\textbf{General NN hyperparameters} \\
\quad Optimizer & Adam & Adam & Adam \\
\quad Learning rate            & $2.8\cdot 10^{-5}$ & $10^{-3}$ & $10^{-3}$ \\ 
\quad Activation function      & tanh                         & tanh                      & tanh                     \\
\quad $L_2$ regularization $\lambda_\text{reg}$ (see Eq.~\eqref{eq: loss})  & $2.8\cdot 10^{-6}$ & $10^{-5}$ & $10^{-6}$ \\
\quad $\lambda_\Phi$ (see Eq.~\eqref{eq: loss_reg})& $1.0$ & $1.0$ & $1.0$ \\
\quad $\lambda_\Psi$ (see Eq.~\eqref{eq: loss_reg})  & $1.0$ & $1.0$ & $1.0$ \\

\midrule
\textbf{\#Trainings for ensemble UQ} $m$ & 30 & 3 & 15 \\
\midrule
{\textbf{Offline training time per epoch}} \\
  \quad {Level 0} &  {0.24 s} & {0.13 s} &  {0.13 s} \\
  \quad {Level 1} &  {0.35 s} &  {0.16 s} & {0.21 s} \\
  \quad {Level 2} &  {0.57 s} & {0.22 s} &  {0.26 s} \\ 
  \quad {Level 3} &  &  {0.32 s} & {0.37 s}\\
\midrule
{\textbf{Online data cost (rel. to HF)}} \\
  \quad {Level 0} &  {0\% (no costs)} & {0\% (no costs)} &  -- \\
  \quad {Level 1} &  {0.04\%} &  {0.02\%} & -- \\
  \quad {Level 2} &  {10.6\%} & {0.27\%} &  -- \\ 
  \quad {Level 3} &  &  {7.27\%} & --\\
\bottomrule
\end{tabular}
\caption{Summary of network architectures, training hyperparameters and computational deaitls for the different applications.}
\label{tab: HPs}
\end{table}
This appendix provides a summary of the implementation, key hyperparameters and network architectures used for the different case studies presented in the main text. 
The specific choices for each application are detailed in Table~\ref{tab: HPs}.

The different \textit{encoder types} reflect the input data modalities of the physical systems, as explained in the \textit{Multi-fidelity setting} Sections \ref{sec: mf-setting RD}-\ref{sec: mf-setting NS}-\ref{sec: mf-setting air}.
For the encoder neural networks (NNs), the input and output dimensions are given by \textit{input dimension $d_\text{in}$} and \textit{latent dimension $d_h$}, respectively. 
When a dimension is reported with $N_\text{POD}$, 
the data are first reduced using POD to the specified number of modes, and the network is trained on the resulting POD coefficients.
The number of hidden layers and nodes for each NN is indicated as a list, where the number of elements corresponds to the number of hidden layers and the values specify the number of nodes per hidden layer.

For the decoders, the input dimension is given by $d^{(l)}_{h_\text{tot}}=\sum_{j\leq l}d_h^{(j)}$ at level $l$, while the \textit{output dimension} is $d_\text{out}$ at every level. The decoder architectures and the number of layers/nodes are indicated under \textit{decoder type} and \textit{decoder NN layers/nodes}, respectively, and are the same at every level since a single high-fidelity quantity is modeled.

Moreover, Table~\ref{tab: HPs} lists general NN hyperparameters used for training. 
The Adam optimizer was used in all cases, with learning rates and $L_2$ regularization coefficients tuned individually.
{All neural networks employ the $\tanh$ activation function, which is well suited for regression tasks involving time-series data, providing smooth and bounded activations that support stable training.}
The regularization weights $\lambda_\Phi$ and $\lambda_\Psi$ for the encoder and decoder networks, respectively, were kept equal to 1.0 in all experiments, with the overall regularization strength $\lambda_\text{reg}$ controlled by the reported value.
\\
We also report, the number of independent retrainings, $m$,  with randomly initialized weights used to compute statistical moments for ensemble-based uncertainty quantification.
{Glorot uniform initialization scheme is used for all network weights, ensuring consistent and unbiased starting points for each ensemble member.}
{The multiple trainings are independent so they can be executed in parallel; otherwise, the computational cost scales linearly with $m$. The ensemble approach provides an implementation-agnostic estimate of epistemic uncertainty, as it does not require modifications to the training procedure or network architecture.
For the Navier–Stokes example, only $m=3$ independent retrainings were performed. This number is insufficient to obtain reliable estimates of higher-order statistical moments, in particular the variance. For this reason, uncertainty measures were not reported for this case, and only ensemble-averaged predictions were shown.}
\paolo{In large-scale or time-sensitive settings, when the computational overhead of ensemble retraining outweighs its simplicity and generality, one may instead adopt probabilistic surrogates such as Bayesian neural networks or conditional neural processes \cite{garnelo2018conditional, wu2022multi}, which provide uncertainty estimates without repeated retrainings.}

{Table~\ref{tab: HPs} also reports indicative computational costs associated with the proposed framework. The \textit{offline training time per epoch} corresponds to the average wall-clock time required to train the neural networks at each fidelity level, measured on the same hardware platform (MacBook Apple M1 8-core processor).}
{The \textit{online data cost (relative to HF)} provides an approximate estimate of the cost required to generate or acquire the input data at each fidelity level during inference, expressed as a percentage of the cost of computing a full high-fidelity solution, which is normalized to 100\%. These estimates are based on the relative number of degrees of freedom of the corresponding data sources. For the reaction-diffusion and Navier-Stokes examples, Level~0 incurs zero online cost, as it only requires control parameters.}
{For the air-pollution case study, online data costs are not reported, as the inputs consist of real sensor data and acquisition and assimilation costs are not provided.}

{Different normalization strategies are adopted depending on the application. For the reaction–diffusion and air-pollution examples, all inputs and outputs are rescaled to the unit interval using max-scaling applied independently to each variable. For the Navier–Stokes example, the HF output POD coefficients are instead normalized by their corresponding singular values. This ensures comparable magnitudes across modes, avoids bias toward high-energy modes during training, and improves the prediction of lower-variance yet dynamically informative components.
Remaining implementation details are available in the accompanying codebase at \url{https://github.com/ContiPaolo/Progressive-Multifidelity-NNs}, ensuring full reproducibility of the results.}

\bibliographystyle{unsrt}
\bibliography{references.bib}

\end{document}